\documentclass[10pt,twocolumn,letterpaper]{article}

\usepackage{cvpr}
\usepackage{times}
\usepackage{epsfig}
\usepackage{graphicx}
\usepackage{amsmath}
\usepackage{amssymb}
\usepackage{xcolor}
\usepackage{booktabs}
\usepackage{float}

\usepackage[pagebackref=true,breaklinks=true,letterpaper=true,colorlinks,bookmarks=false]{hyperref}

\cvprfinalcopy % *** Uncomment this line for the final submission

\def\cvprPaperID{****} % *** Enter the CVPR Paper ID here

\ifcvprfinal\fi
\begin{document}

%%%%%%%%% TITLE
\title{Understanding Textual Emotion Through Emoji Prediction}
\author{
Ethan Gordon \quad Nishank Kuppa \quad Rigved Tummala \quad Sriram Anasuri\\
College of Computing, Georgia Institute of Technology\\
Atlanta, GA 30332\\
{\tt\small \{egordon40, nkuppa3, rtummala6, sanasuri3\}@gatech.edu}
}

\maketitle
%\thispagestyle{empty}

%%%%%%%%% ABSTRACT
\begin{abstract}
\vspace{-10pt}
   % The ABSTRACT is to be in fully-justified italicized text, at the top
   % of the left-hand column, below the author and affiliation
   % information. Use the word ``Abstract'' as the title, in 12-point
   % Times, boldface type, centered relative to the column, initially
   % capitalized. The abstract is to be in 10-point, single-spaced type.
   % Leave two blank lines after the Abstract, then begin the main text.
   % Look at previous CVPR abstracts to get a feel for style and length. 
   % The abstract section should contain a brief summary of your work that
   % includes the problem statement, proposed solution and results.

This project explores emoji prediction from short text sequences using four deep learning architectures: a feedforward network, CNN, transformer, and BERT. Using the TweetEval dataset, we address class imbalance through focal loss and regularization techniques. Results show BERT achieves the highest overall performance due to it's pretraining advantage, while CNN demonstrates superior efficacy on rare emoji classes. This research shows the importance of architecture selection and hyperparameter tuning for sentiment-aware emoji prediction, contributing to improved human-computer interaction.

\end{abstract}
\vspace{-15pt}
%%%%%%%%% BODY TEXT
\section{Introduction/Background/Motivation}

% \textbf{(5 points) What did you try to do? What problem did you try to solve? Articulate your objectives using absolutely no jargon. }

\subsection{Problem Statement and Objectives}
\vspace{-5pt}
This project entails building various deep learning models that can effectively predict which emoji best matches a short text message. This task is closely related to sentiment analysis, where the goal is to detect the emotion or mood based on a piece of text. 

The primary goal of this project is to train machine learning models to understand the direct or indirect meaning behind a sequence of words and then select the emoji that best fits it. Some emojis are used very often (such as heart emojis), while others are more occasional (such as Christmas tree emojis), which makes it harder for the model to learn to predict the less common ones. The challenge is therefore to build a model that can go beyond just picking the most frequent emoji and instead learn to match the right emoji to each unique message to express the correct sentiment. 
The objectives of this project are:
\begin{enumerate}
    \item To build models that can match short messages to emojis in a way that feels accurate and human-like.
    \vspace{-5pt}
    \item To improve predictions for rare emojis that are not frequently used.
    \vspace{-5pt}
    \item To compare different model designs and hyperparameter tuning strategies to find what works best.
\end{enumerate}

\subsection{Current Methods and Limitations}
\vspace{-5pt}
% \textbf{(5 points) How is it done today, and what are the limits of current practice?}

Emoji sentiment classification is currently performed in various ways: lexicon-based methods, machine learning models, and transformer-based models. 
Lexicon-based methods, which assign predefined sentiment scores to a dataset of emojis, are rudimentary and simple compared to machine learning methods. They essentially behave like keyword lookup systems, where words are matched to emojis. In recent years, classical machine learning models (such as SVMs and  Naïve Bayes), and deep learning models (such as CNNs and transformers) have been used by training them on large datasets \cite{Kulshreshtha2025}. In mobile keyboard emoji prediction, a lightweight machine learning model, such as an RNN, that operates directly on the device is used \cite{beaufays2019federated}.

Even with these advanced methods, many issues still persist. These models often fail to handle context accurately, especially when sarcasm or cultural variation is involved. For instance, the crying emoji has recently been used on social media to express uncontrollable laughter instead of sadness, so models need to be able to keep up with multiple meanings of the same emoji. In general, the models tend to overfit to frequent patterns, and older lexicon-based methods cannot adequately keep up with how emoji usage evolves. To summarize, current methods still fall short when it comes to nuance and actual human-like expression, and this project aims to address these issues.

\subsection{Impact}
\vspace{-5pt}
% \textbf{(5 points) Who cares? If you are successful, what difference will it make? }

% The impact of this project spans from end users who want better emoji suggestions in their smartphone keyboards to developers who build messaging apps or social media platforms. If emoji prediction becomes more accurate and context-aware, users will receive better emoji suggestions that feel more realistic and expressive. Better emoji prediction also means better adaptability to new emoji usage patterns and a more personalized experience for the user. Beyond convenience, improved emoji prediction enhances a machine’s ability to understand human emotion in text, which makes interactions feel less robotic and more personal. Success here means bridging the gap between language and emotion, which is valuable for more human-centered AI.
The impact of this project spans from helping users get more accurate and expressive emoji suggestions to supporting developers of messaging apps and social media platforms that rely on understanding user content. If emoji prediction becomes more accurate and context-aware, it can improve user experience through smarter suggestions and help platforms better interpret user sentiment for content moderation or recommendations. It also allows systems to adapt to new emoji trends and provide and a more personalized experience for the user across digital communication platforms. Beyond the convenience, this research shows that emoji prediction serves as an effective testbed for evaluating sentiment analysis architectures, providing insights that extend to broader natural language processing applications where more emotional nuance matters. Success in this project means closing the gap between language and emotion while advancing our understanding of how different deep learning architectures handle precise sentiment classification, which is valuable for more human-centered AI.    

\subsection{Dataset Selection}
\vspace{-5pt}
% \textbf{(5 points) What data did you use? Provide details about your data, specifically choose the most important aspects of your data mentioned \href{https://arxiv.org/abs/1803.09010}{here}. You don’t have to choose all of them, just the most relevant.}

This project uses the \href{https://huggingface.co/datasets/cardiffnlp/tweet_eval}{TweetEval emoji prediction dataset} by Barbieri et al. from HuggingFace. The dataset simply consists of two columns: the first column contains a tweet, and the second containing a class label represented by one of 20 emojis that reflect the sentiment of the tweet. Tweets are a good text representation for this project because they are short, informal, and often have direct emotional cues, which makes them ideal for studying how people pair language with emojis in normal communication.

The dataset is based on publicly available tweets and no further preprocessing is required, since the structure of the data is already straightforward and well-formatted. The emoji configuration of the dataset, which is used for this project, consists of 45,000 training samples, 5,000 validation samples, and 50,000 test samples. The models presented in this paper are trained using all of the available training samples. 

The dataset contains class imbalance since some emojis, such as the red heart, appear far more often than others. While this poses a challenge for accurately predicting rarer emojis, the dataset represents real-world and informal language that provides a variety of data for studying common emoji usage patterns.

%-------------------------------------------------------------------------
%------------------------------------------------------------------------
\section{Approach}

% \textbf{(10 points) What did you do exactly? How did you solve the problem? Why did you think it would be successful? Is anything new in your approach? }
\subsection{Design Choices and Implementation}
\vspace{-5pt}

To conduct this project and explore various approaches, four models are created using different architectures: BERT, a feedforward neural network, a transformer, and a CNN. This approach of evaluating four vastly different architectures and comparing their performance is novel for many reasons. Firstly, these models are chosen to compare the different strengths in each model: BERT for its transfer learning from large-scale pretraining, the feedforward network as a baseline, the CNN for learning localized semantic patterns, and the transformer for modeling dependencies between tokens that may not necessarily be near each other using attention mechanism. Secondly, this project aims to explore the class imbalance due to certain emojis dominating the dataset, and it is worthwhile seeing which model can handle the imbalance the best and more accurately predict the rarer emojis. It should be noted the custom models like feedforward/transformer/CNN leverage the pytorch framework and a TweetTokenizer to tokenize our words. 

All four architectures have the same initial setup:
\begin{enumerate}
    \item \textbf{Dataset Loading:} After importing relevant libraries (such as torch, torchtext, datasets, and nltk), the TweetEval dataset with the "emoji" configuration is loaded and split into training, validation, and test sets.
    \vspace{-5pt}
    \item \textbf{Tokenization and Building Vocabulary:} A pretrained tokenizer from nltk, called TweetTokenizer, is used to tokenized the tweets. This tokenizer is optimized for tweets. Afterwards, a vocabulary is built based on the training data and special tokens for padding and unknown words are included.
    \vspace{-5pt}
    \item \textbf{Encoding:} The tweet text is encoded into token IDs, and padding or truncation is applied to conform to the fixed length of 64. 
    \vspace{-5pt}
    \item \textbf{DataLoaders:} DataLoaders are created for all of the sets to help with batching during training and test.
\end{enumerate}

Below is a more detailed overview of each architecture. Justifications will be provided in Section 3: Experiments and Results.
\begin{itemize}
    \item \textbf{BERT:}
    We use a pre-trained BERTweet base model to get contextualized hidden representations with 768 dimensions for each token. We then run these through three different attention mechanisms: word-level attention with 8 heads, phrase-level attention with 4 heads, and sentence-level attention with 2 heads. A 1D convolutional layer with kernel size 3 and adaptive average pooling was added. Each attention stream goes through masked pooling to handle different sequence lengths and ignore padding. We combine all four streams and use a fusion layer to bring the 3072 dimensions back down to 768. After applying layer normalization, ReLU, and dropout, we feed everything through a two-layer classifier that goes from 768 to 384 to the final emoji classes, with ReLU and dropout.
    \vspace{-5pt}
    \item \textbf{Feedforward Network:} This network first maps input tokens to dense vectors using an embedding layer. Max pooling is applied across the sequence to extract the most salient features, reducing the input to fixed-size feature vectors. This pooled vector is passed through three linear layers that go from 256 to 128 to 64 dimensions. ReLU activations, layer normalization, and dropout for regularization are used. The final output layer produces logits over the target emoji classes. 
    \vspace{-10pt}
    \item \textbf{Transformer:} This model starts similar to the feedforward network with it's embedding layer, along with providing positional encodings to retain sequential order of the tokens. These embeddings are then passed through a multi-layer Transformer encoder, which applies self-attention to model possible relationships between tokens. We opted to have 2 transformer encoder layers with 2 attention heads each. After the Transformer, we apply a max pooling operation across the token dimension to obtain a fixed-length output and aggregate strongest activation per token. These results are then passed through a fully connected classifier with an input of 128 dimensions as well as a ReLU activation to produce the final emoji classification.
    \vspace{-5pt}
    \item \textbf{CNN:} The proposed Convolutional Neural Network employs a multi-kernel architecture. It is designed to capture different n-gram patterns essential for sentiment-based emoji classification. This implementation uses three parallel convolutional layers with kernel sizes of 3, 4, and 5, which allows for trigram, 4-gram, 5-gram feature extraction, and this correlates to the different levels of semantic granularity with the tweet text. Applying global max pooling to each convolutional output ensuring translation invariance for emotional expressions. The key hyperparameters include 128 dimensional embeddings, 128 filters per layer, 0.3 dropout, and focal loss with balanced class weights (gamma=1.5) to address class imbalance. 
\end{itemize}

% \textbf{(5 points) What problems did you anticipate? What problems did you encounter? Did the very first thing you tried work? }
\subsection{Problems Anticipated and Encountered}

% \textbf{Important: Mention any code repositories (with citations) or other sources that you used, and specifically what changes you made to them for your project. }

We expected a challenge with varying amounts of tokens in tweets (i.e different length tweets). To overcome this challenge, we simply add padding tokens to uniformly match input data for our models, particularly when dealing with batched data. Going into this problem we also anticipated limited accuracy based on domain knowledge and previous attempts in other works, as tweets often have informal dialogue, inaccurate use of words, and changing sentence context based on culture changes. We also have a major class imbalance favoring the :heart: emoji with 10,000+ instances, and some other ones like :heart\_eyes: and :joy: around 4,500+ instances, compared to that of something like :grin: which only occurs 1153 times. As such, we originally tried regular cross entropy loss, but found that it gave less-than-desirable results, and saw much better performance/gradient flow with focal loss.

\section{Experiments and Results}

% \textbf{(10 points) How did you measure success? What experiments were used? What were the results, both quantitative and qualitative? Did you succeed? Did you fail? Why? Justify your reasons with arguments supported by evidence and data.}

% \textbf{Important: This section should be rigorous and thorough. Present detailed information about decision you made, why you made them, and any evidence/experimentation to back them up. This is especially true if you leveraged existing architectures, pre-trained models, and code (i.e. do not just show results of fine-tuning a pre-trained model without any analysis, claims/evidence, and conclusions, as that tends to not make a strong project). }
\vspace{-5pt}
Success for each architecture is measured using standard classification metrics, namely:
\begin{itemize}
    \item \textbf{Accuracy:} The overall percentage of correctly classified tweets. While a straightforward metric, this can be misleading in datasets with class imbalance, such as the TweetEval dataset.
    \vspace{-5pt}
    \item \textbf{Loss:} Both training and validation losses are tracked. Importantly, focal loss is used for validation, since it is designed to handle class imbalance by focusing on more difficult examples.
    \vspace{-5pt}
    \item \textbf{Precision, Recall, and F1-score:} These metrics provide a more specific view of performance per class.
    % \begin{itemize}
    %     \item \textbf{Precision:} Of all the tweets predicted as a certain emoji, how many actually were that emoji?
    %     \item \textbf{Recall:} Of all the tweets that should have been classified as a certain emoji, how many were correctly identified?
    %     \item \textbf{F1-score:} The harmonic mean of precision and recall. A single metric that balances both.
    % \end{itemize}
    
\end{itemize}

% \begin{wrapfig}
\subsection{BERT}
Our multi-scale attention model uses various tuned hyperparameters. We experimented with higher dropout values but didn't find much benefit, so we settled on 0.3 for dropout in the fusion layer and 0.2 for the classifier to help reduce overfitting. We utilized pre-trained 768-dimensional BERTweet embeddings specifically designed for social media text with a batch size of 16 due to memory constraints. We used the AdamW optimizer with a conservative learning rate of 2e-5 and weight decay of 0.01, since AdamW provided better regularization than standard Adam for fine-tuning pre-trained models. We experimented with higher weight decay values and different learning rates, but the variations showed little improvement, so we stuck with our conservative approach. Finally, we decided on 3 epochs with gradient clipping at 1.0, since any further epochs would lead to severe overfitting.
The multi-scale attention architecture captures different linguistic patterns at varying granularities. Word-level attention with 8 heads focuses on fine-grained token relationships and requires more attention capacity. Phrase-level attention with 4 heads operates on local chunks, identifying word combinations that indicate emoji usage patterns. Sentence-level attention with 2 heads captures broader understanding for overall sentiment and long-range dependencies. The 1D convolutional layer with kernel size 3 extracts complementary features that attention mechanisms might miss, particularly useful for sequential patterns in social media text. This resulted in a well-balanced architecture handling multiple levels of linguistic complexity. Our model evaluation focused on both accuracy and weighted F-1 score to account for severe class imbalance. The final model achieved 44 percent accuracy with a weighted F-1 score of 0.45, a significant improvement over baseline approaches. The model excelled on emojis with distinctive patterns: heart emoji (F-1: 0.81), Christmas tree (F-1=0.71), and American flag emoji (F-1=0.62), showing effective capture of emotional, seasonal, and political context markers. However, class imbalance remained challenging, with rare classes like winking tongue emoji (F-1=0.11) and grinning emoji (F-1=0.11) being largely overlooked. This shows that while the architecture picked up on common emotional patterns, it struggled with subtle contextual differences in less frequent emojis. Our learning curve shows training loss decreasing from 2.43 to 1.77 over 3 epochs, while validation loss stabilizes around 2.32 after epoch 1. The gap between training and validation loss suggests mild overfitting, but much more controlled than typical transformer architectures.

%\centering
%\fbox{\includegraphics[width=0.9\linewidth]{image.png}}
%\caption{Learning Curve for Bert}
%\label{fig:Learning Curve for BERT}
%\end{wrapfig}
\begin{figure}[htbp]
    \centering
    \fbox{\includegraphics[width=0.75\linewidth]{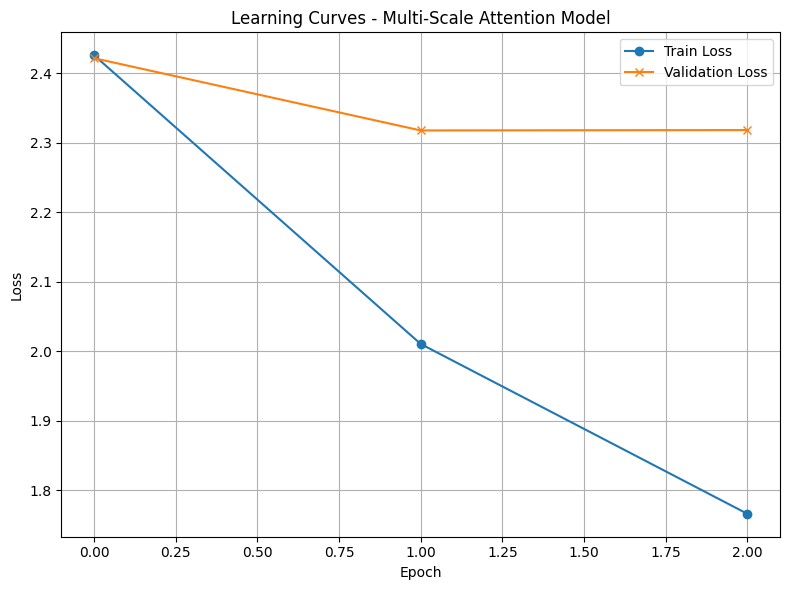}}
    \caption{Learning Curve for BERT}
    \label{fig:Learning Curve for BERT Network}
\end{figure}

\begin{table}[htbp]
\centering
\resizebox{\columnwidth}{!}{%
\begin{tabular}{lcccc}
\toprule
\textbf{Label} & \textbf{Precision} & \textbf{Recall} & \textbf{F1-Score} & \textbf{Support} \\
\midrule
:heart:           & 0.91 & 0.73 & 0.81 & 10798 \\
:heart\_eyes:     & 0.44 & 0.24 & 0.31 & 4830 \\
:joy:             & 0.56 & 0.43 & 0.49 & 4534 \\
:two\_hearts:     & 0.21 & 0.15 & 0.18 & 2605 \\
:fire:            & 0.62 & 0.53 & 0.57 & 3716 \\
:blush:           & 0.16 & 0.17 & 0.17 & 1613 \\
:sunglasses:      & 0.23 & 0.20 & 0.21 & 1996 \\
:sparkles:        & 0.32 & 0.36 & 0.34 & 2749 \\
:blue\_heart:     & 0.17 & 0.23 & 0.19 & 1549 \\
:kiss:            & 0.15 & 0.32 & 0.20 & 1175 \\
:camera:          & 0.33 & 0.51 & 0.40 & 1432 \\
:flag-us:         & 0.58 & 0.68 & 0.62 & 1949 \\
:sunny:           & 0.43 & 0.81 & 0.56 & 1265 \\
:purple\_heart:   & 0.11 & 0.17 & 0.14 & 1114 \\
:wink:            & 0.13 & 0.18 & 0.15 & 1306 \\
:100:             & 0.24 & 0.37 & 0.29 & 1244 \\
:grin:            & 0.10 & 0.12 & 0.11 & 1153 \\
:christmas\_tree: & 0.65 & 0.79 & 0.71 & 1545 \\
:camera\_with\_flash: & 0.37 & 0.27 & 0.31 & 2417 \\
:stuck\_out\_tongue\_winking\_eye: & 0.09 & 0.13 & 0.11 & 1010 \\
\midrule
\textbf{Accuracy}     &       &       & 0.44 & 50000 \\
\textbf{Macro Avg}    & 0.34  & 0.37  & 0.34 & 50000 \\
\textbf{Weighted Avg} & 0.48  & 0.44  & 0.45 & 50000 \\
\bottomrule
\end{tabular}
}
\caption{\centering Emoji Classification Report (BERT Model)}
\label{tab:emoji_classification_bert}
\end{table}

% \begin{wrapfig}
\subsection{Feedforward Network}
\begin{figure}[H]
    \centering
    \fbox{\includegraphics[width=0.75\linewidth]{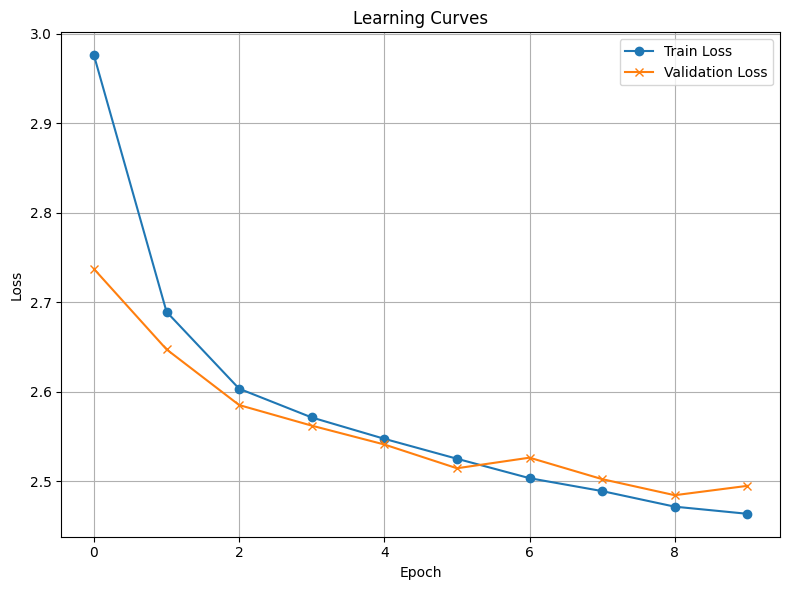}}
    \caption{Learning Curve for Feedforward Network}
    \label{fig:Learning Curve for Feedforward Network}
\end{figure}

Regarding the neural network architecture, three linear layers that go from 256 to 128 to 64 dimensions proved to be the best setup, as more layers led to overfitting and fewer layers led to underfitting. Max pooling also proves to be useful by extracting the most informative features across the sequence, thereby effectively highlighting the most important words in each tweet.
Regarding hyperparameters, an embedding dimension of 128 is used to represent each token, which provides enough expressiveness for the tweets. Dropout is set to 0.3 to reduce overfitting (experimentation showed that this value works well, while lower values were minimal in benefit). A batch size of 32 is used during training, and the Adam optimizer is applied with a small learning rate of 5e-4 and weight decay of 1e-4 to stabilize learning, improve gradient updates, and minimize overfitting. The model trains for 10 epochs, as experimentation showed that any further epochs lead to severe overfitting. 
Figure \ref{fig:Learning Curve for Feedforward Network} shows the learning curve for the feedforward network. The curve indicates that although overfitting begins at epoch 6, it is mild in nature. Overall, the validation loss stays close to the training loss and both curve trend downwards, so the learning process is stable. Experimentation shows that using focal loss was instrumental in achieving a good learning curve like this, as using class-weighted cross entropy loss led to a much larger gap between the training loss and validation loss. This is likely because focal loss is better for datasets that have a skewed distribution and multiple rare classes like this one.
Table \ref{tab:emoji_classification_feedforward} presents the classification report for this neural network. The report indicates that the network achieves an overall accuracy of only 28\% and a weighted F1-score of 0.28. The class imbalance is apparent, as the :heart: class is heavily overpredicted by the model. Many rare classes, such as 13, 15, and 16, are nearly ignored, with F1-scores close to zero. Only a few mid-frequency classes perform moderately well, and class 17 stands out with an F1-score of 0.61. Although focal loss is used to mitigate the class imbalance, other techniques such as random oversampling may further mitigate the imbalance.
%\centering
%\fbox{\includegraphics[width=0.9\linewidth]{learning_curve_FF.png}}
%\caption{Learning Curve for Feedforward Network}
%\label{fig:Learning Curve for Feedforward Network}
%\end{wrapfig}

\begin{table}[htbp]
\centering
\resizebox{\columnwidth}{!}{%
\begin{tabular}{lcccc}
\toprule
\textbf{Label} & \textbf{Precision} & \textbf{Recall} & \textbf{F1-Score} & \textbf{Support} \\
\midrule
:heart:           & 0.95 & 0.50 & 0.66 & 10798 \\
:heart\_eyes:     & 0.19 & 0.08 & 0.11 & 4830 \\
:joy:             & 0.21 & 0.45 & 0.29 & 4534 \\
:two\_hearts:     & 0.17 & 0.18 & 0.17 & 2605 \\
:fire:            & 0.22 & 0.37 & 0.28 & 3716 \\
:blush:           & 0.08 & 0.02 & 0.03 & 1613 \\
:sunglasses:      & 0.08 & 0.18 & 0.11 & 1996 \\
:sparkles:        & 0.09 & 0.02 & 0.04 & 2749 \\
:blue\_heart:     & 0.11 & 0.03 & 0.04 & 1549 \\
:kiss:            & 0.10 & 0.21 & 0.13 & 1175 \\
:camera:          & 0.26 & 0.17 & 0.20 & 1432 \\
:flag-us:         & 0.29 & 0.12 & 0.17 & 1949 \\
:sunny:           & 0.21 & 0.67 & 0.32 & 1265 \\
:purple\_heart:   & 0.00 & 0.00 & 0.00 & 1114 \\
:wink:            & 0.07 & 0.09 & 0.08 & 1306 \\
:100:             & 0.02 & 0.00 & 0.00 & 1244 \\
:grin:            & 0.05 & 0.00 & 0.00 & 1153 \\
:christmas\_tree: & 0.57 & 0.64 & 0.61 & 1545 \\
:camera\_with\_flash: & 0.36 & 0.37 & 0.37 & 2417 \\
:stuck\_out\_tongue\_winking\_eye: & 0.04 & 0.13 & 0.06 & 1010 \\
\midrule
\textbf{Accuracy}     &       &       & 0.28 & 50000 \\
\textbf{Macro Avg}    & 0.20  & 0.21  & 0.18 & 50000 \\
\textbf{Weighted Avg} & 0.35  & 0.28  & 0.28 & 50000 \\
\bottomrule
\end{tabular}
}
\caption{\centering Emoji Classification Report (Feedforward Network)}
\label{tab:emoji_classification_feedforward}
\end{table}

\subsection{Transformer}
The transformer model uses various tuned hyper parameters. We decided on using 0.3 for dropout to try to combat overfitting, which we will discuss shortly; and didn't find much benefit in higher dropout values. Similar to our feedforward network, we opted for an embedding dimension of 128 to give our tokens sufficient feature representation. Similar to feedforward, we chose a batch size of 32. To further combat overfitting we utilized the AdamW optimizer along with a moderate learning rate of 1e-4 and a weight decay of 4e-5. Marginally better results were found with the AdamW optimizer versus the Adam optimizer since the weight updates are detached from the gradients-leading to better generalization. Higher weight decay values were experimented with along with lower/high learning rates, but variation showed little improvement to combat overfitting. Lastly, we decided on 15 epochs, but included an early stopping check on validation data loss to attempt to reduce overfitting in the training process.

From Table \ref{tab:emoji_classification_transformer_updated}, we see small improvement in overall accuracy comparing between 0.28 and 0.30, and slightly better macro and weighted F1-score indicating better performance on certain minority classes. Notably, classes such as :purple\_heart:, :fire:, :camera:, and :flag-us: show marked increases in precision and recall, with some classes like :purple\_heart: moving from zero to measurable F1-scores. However, these gains often come with trade-offs, as some classes (like :joy: and :two\_hearts:) experience decreases in recall and F1-score. Overall, while switching from a feedforward to a transformer encoding architecture does yield some class-specific improvements, the architectural change alone does not produce a dramatic boost in overall performance after tuning. The main impact is improved representation of previously under performing classes that received an F1-score of 0, but often at a slight cost to other classes.

\begin{figure}[H]
    \centering
    \fbox{\includegraphics[width=0.75\linewidth]{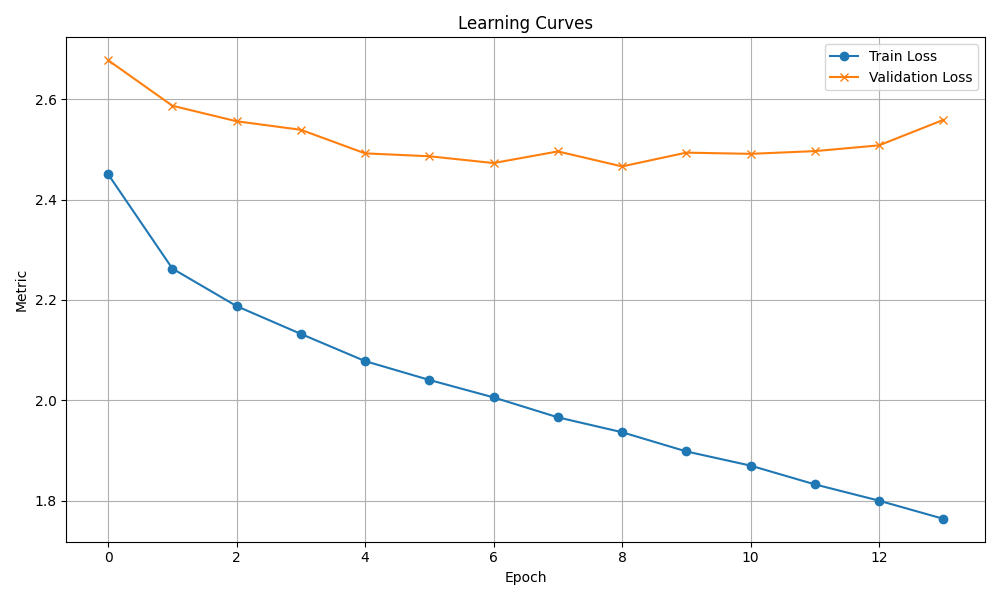}}
    \caption{Learning Curve for Transformer Network}
    \label{fig:Learning Curve for Transformer Network}
\end{figure}

Lastly, we look at our learning curve showing our loss over epochs for training/validation data as shown in Figure \ref{fig:Learning Curve for Transformer Network}. We see that we have lower loss than that of our feedforward network, but we have a gap between our training/validation loss that continues to worsen over epochs. This suggests that our model is overfitting as validation loss remains steadfast and training loss continues to decrease. Despite extensive hyperparameter tuning to reduce model complexity (fewer transformer layers, smaller embedding dimensions, fewer attention heads) and adjusting learning rate and weight decay; the generalization gap remained. Nevertheless, the transformer still delivers strong performance. With better regularization or data augmentation to counteract overfitting and class imbalance, it’s likely the transformer could outperform the feedforward network more substantially.

\begin{table}[H]
\centering
\resizebox{\columnwidth}{!}{%
\begin{tabular}{lcccc}
\toprule
\textbf{Label} & \textbf{Precision} & \textbf{Recall} & \textbf{F1-Score} & \textbf{Support} \\
\midrule
:heart:           & 0.92 & 0.66 & 0.77 & 10798 \\
:heart\_eyes:     & 0.18 & 0.11 & 0.14 & 4830 \\
:joy:             & 0.25 & 0.31 & 0.28 & 4534 \\
:two\_hearts:     & 0.17 & 0.06 & 0.09 & 2605 \\
:fire:            & 0.46 & 0.28 & 0.35 & 3716 \\
:blush:           & 0.09 & 0.03 & 0.05 & 1613 \\
:sunglasses:      & 0.17 & 0.08 & 0.10 & 1996 \\
:sparkles:        & 0.20 & 0.12 & 0.15 & 2749 \\
:blue\_heart:     & 0.09 & 0.11 & 0.10 & 1549 \\
:kiss:            & 0.09 & 0.25 & 0.13 & 1175 \\
:camera:          & 0.25 & 0.49 & 0.33 & 1432 \\
:flag-us:         & 0.47 & 0.35 & 0.40 & 1949 \\
:sunny:           & 0.26 & 0.43 & 0.33 & 1265 \\
:purple\_heart:   & 0.05 & 0.13 & 0.08 & 1114 \\
:wink:            & 0.06 & 0.24 & 0.10 & 1306 \\
:100:             & 0.08 & 0.08 & 0.08 & 1244 \\
:grin:            & 0.05 & 0.10 & 0.07 & 1153 \\
:christmas\_tree: & 0.51 & 0.70 & 0.59 & 1545 \\
:camera\_with\_flash: & 0.33 & 0.07 & 0.12 & 2417 \\
:stuck\_out\_tongue\_winking\_eye: & 0.04 & 0.08 & 0.05 & 1010 \\
\midrule
\textbf{Accuracy}     &       &       & 0.30 & 50000 \\
\textbf{Macro Avg}    & 0.24  & 0.23  & 0.21 & 50000 \\
\textbf{Weighted Avg} & 0.38  & 0.30  & 0.32 & 50000 \\
\bottomrule
\end{tabular}%
}
\caption{\centering Emoji Classification Report (Transformer Model)}
\label{tab:emoji_classification_transformer_updated}
\end{table}

\subsection{Convolutional Neural Network} Initially overfitting with a test accuracy of 35\% and a training accurarcy of 72\%, the model was redesigned in order to achieve more balanced results and healthier training patterns. The tuning process included speeding up the learning rate from 1e-4 to 1e-3, epochs was increased to 20, but early stopping was implemented with a patience of 3 (earlier stoppage than default patience of 5). However, the accuracy would peak on the validation data and then drop after 5 epochs while training accuracy increased rapidly. But due to early stopping the run was getting cut off around 8 epochs, so it made sense to drop epochs down to 5 since the model was learning too fast and much that it was starting to overfit after that point. This resulted in a very balanced final accuracy of 31.35\% (training) and 32.7\% (test).

\begin{figure}[htbp]
    \centering
    \fbox{\includegraphics[width=0.75\linewidth]{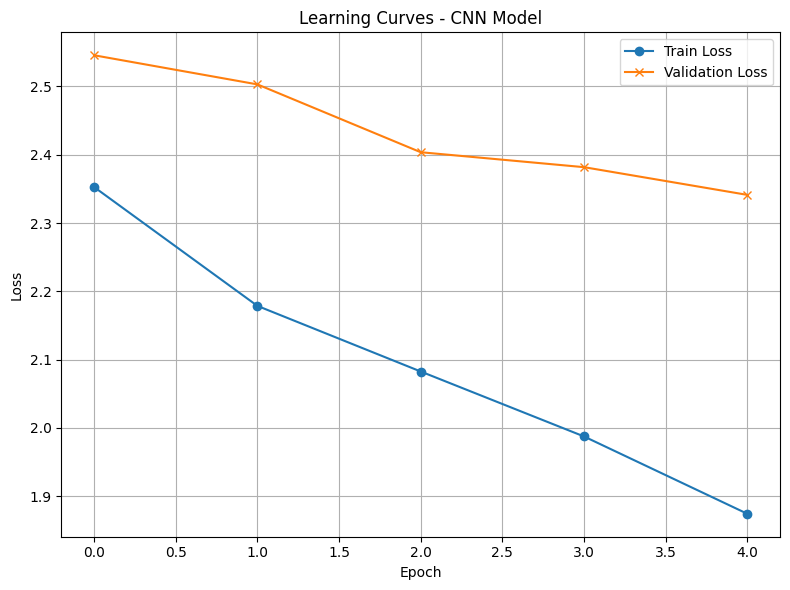}}
    \caption{Learning Curve for Convolutional Neural Network}
    \label{fig: Learning Curve for Convolutional Neural Network}
\end{figure}

Our model success while including accuracy also was focused on weighted F-1 score in order to account for class imbalance. The final CNN had an overall accuracy 33\% with a weighted F-1 score of 0.34, which was a subtle improvement over the transformer and the feedforward baseline. The learning curve [\ref{fig: Learning Curve for Convolutional Neural Network}] shows that the model undergoes stable learning with consistent training loss across the five epochs. There is a very small gap after the second epoch between the two curves, suggesting overfitting, but both curves do trend downward, validating the effectiveness of the regularization strategy.
\vspace{-3pt}

A deep dive into the results per class [\ref{tab:emoji_classification_convolutional}] gives us insight into the model's distinct performance patterns, telling us more about the model's capabilities and limitations. The CNN did exceptionally well on predicting emojis that contain distinctive lexical patterns - Christmas tree got an F-1 score of 0.64 showing the model captured seasonal terminology, the American flag emoji got 0.46 indicating strong political context understanding, and the fire emoji did really well with 0.43 by picking up on modern slang usage. Class imbalance remained a consistent issue, with 22\% of the dataset's samples including the standard heart emoji, leading the model to be very heavily biased to this emoji (F-1=0.75). This imbalance led to issues with semantically similar emojis such as the purple heart (F-1=0.09), the double heart (F-1=0.10), and the blue heart (F-1=0.11), telling us that while the model was able to pick up on emotional sentiment it struggled with subtle contextual understanding. The imbalance also affected less common emoji predictions despite using focal loss, with classes like the winking tongue emoji and the blushing emoji (both F-1=0.07) nearly ignored due to lack of training samples. This gap in performance is reflected in the macro averaged F-1 score of 0.24, which falls well below the weighted average of 0.34. These results show that while convolutional architectures do well at predicting emojis with clear linguistic signals, they still struggle with capturing subtle emotional differences that need deeper contextual understanding beyond just local n-gram patterns.

\begin{table}[htbp]
\centering
\resizebox{\columnwidth}{!}{%
\begin{tabular}{lcccc}
\toprule
\textbf{Label} & \textbf{Precision} & \textbf{Recall} & \textbf{F1-Score} & \textbf{Support} \\
\midrule
:heart:           & 0.96 & 0.62 & 0.75 & 10798 \\
:heart\_eyes:     & 0.27 & 0.06 & 0.10 & 4830 \\
:joy:             & 0.37 & 0.26 & 0.31 & 4534 \\
:two\_hearts:     & 0.20 & 0.07 & 0.10 & 2605 \\
:fire:            & 0.38 & 0.48 & 0.43 & 3716 \\
:blush:           & 0.10 & 0.06 & 0.07 & 1613 \\
:sunglasses:      & 0.10 & 0.18 & 0.13 & 1996 \\
:sparkles:        & 0.25 & 0.19 & 0.21 & 2749 \\
:blue\_heart:     & 0.10 & 0.13 & 0.11 & 1549 \\
:kiss:            & 0.12 & 0.21 & 0.15 & 1175 \\
:camera:          & 0.23 & 0.62 & 0.34 & 1432 \\
:flag-us:         & 0.42 & 0.50 & 0.46 & 1949 \\
:sunny:           & 0.30 & 0.68 & 0.42 & 1265 \\
:purple\_heart:   & 0.06 & 0.18 & 0.09 & 1114 \\
:wink:            & 0.07 & 0.14 & 0.10 & 1306 \\
:100:             & 0.12 & 0.16 & 0.13 & 1244 \\
:grin:            & 0.09 & 0.13 & 0.10 & 1153 \\
:christmas\_tree: & 0.56 & 0.76 & 0.64 & 1545 \\
:camera\_with\_flash: & 0.27 & 0.05 & 0.09 & 2417 \\
:stuck\_out\_tongue\_winking\_eye: & 0.05 & 0.09 & 0.07 & 1010 \\
\midrule
\textbf{Accuracy}     &       &       & 0.33 & 50000 \\
\textbf{Macro Avg}    & 0.25  & 0.28  & 0.24 & 50000 \\
\textbf{Weighted Avg} & 0.40  & 0.33  & 0.34 & 50000 \\
\bottomrule
\end{tabular}
}
\caption{\centering Emoji Classification Report (Convolutional Network)}
\label{tab:emoji_classification_convolutional}
\end{table}

\section{Conclusion}
This comprehensive study analyzed four different deep learning architectures for emoji prediction from a given text, giving insight into the relationship between design of the model and performance on sentiment classification tasks. BERT emerged as the clear top performer among the four models with an accuracy of 44\% and a weighted F-1 score of 0.45. The design employs sophisticated multi-scale attention mechanism and pre-trained social media representations to excel on emojis with distinctive patterns like heart (F1=0.81) and Christmas tree (F1=0.71). Throughout our experimentation another consistent problem all the models ran into was class imbalance due to the heart emoji's dominance in terms of sample count within the dataset used, creating bias regardless of our implemented offsets in terms of focal loss or class weighting strategies. All the models demonstrated consistent strong performance on emojis with clear distinct lexical patterns and poor performance on semantically similar variant emojis (ex. the different heart variants), highlighting the limitations in current approaches to sharper sentiment classification. These results and analysis can be used for HCI and UX/UI implementations, from things like smartphone keyboard improvement to social media content understanding, with the BERT architecture showing the most promise for practical usage with CNN closely following behind it. The research proves that emoji prediction serves as an effective testing method for evaluating model architecture in sentiment analysis, showing clear evidence that the design must align with task characteristics and data properties to achieve optimal performance. Future work could explore better data augmentation, contrastive learning, and hybrid models that combine different strengths to address persistent problems with rare classes and semantic similarity.

% %-------------------------------------------------------------------------
% \section{Other Sections}
\begin{table*}[htbp]
\begin{center}
\begin{tabular}{|l|c|p{8cm}|}
\hline
Student Name & Contributed Aspects & Details \\
\hline\hline
Nishank Kuppa & Feedforward Network Implementation & Identified TweetTokenizer to implement for tokenization. Implemented and tuned deep feedforward network.\\
\newline
Ethan Gordon & Transformer Architecture and Analysis & Implemented a transformer encoder architecture. Analyzed results with a variety of hyperparameters \\
\newline
Rigved Tummala & Convolutional Neural Network and Analysis & Designed and implemented CNN model. Analyzed results while tuning hyperparams for the best performance.\\
\newline
Sriram Anasuri & BERT Model and Analysis & Designed and implemented BERT model. Analyzed results while tuning hyperparams for the best performance.\\
\hline
\end{tabular}
\end{center}
\caption{Contributions of team members.}
\label{tab:contributions}
\end{table*}

\newpage

{\small
\bibliographystyle{ieee_fullname}
\bibliography{MAINREPORT}
}

\end{document}